%% file: emnlp2023.tex
\pdfoutput=1

\documentclass[11pt]{article}

\usepackage[]{EMNLP2023}

\usepackage{times}
\usepackage{latexsym}
\usepackage{multirow}
\usepackage{hyperref}
\usepackage[T1]{fontenc}

\usepackage[utf8]{inputenc}

\usepackage{microtype}

\usepackage{booktabs}

\usepackage[normalem]{ulem}

\usepackage{inconsolata}
\usepackage{enumitem}
\usepackage{graphicx}

\title{Catchy name}

\author{Md Mushfiqur Rahman, Fardin Ahsan Sakib, Fahim Faisal, Antonios Anastasopoulos \\
  Department of Computer Science, George Mason University \\
  \texttt{\{mrahma45,fsakib,ffaisal,antonis\}@gmu.edu }
  }

\title{To token or not to token: A Comparative Study of Text Representations for Cross-Lingual Transfer}
\begin{document}
\maketitle

\begin{abstract}


Choosing an appropriate tokenization scheme is often a bottleneck in low-resource cross-lingual transfer. To understand the downstream implications of text representation choices, we perform a comparative analysis on language models having diverse text representation modalities including 2 segmentation-based models (\texttt{BERT}, \texttt{mBERT}), 1 image-based model (\texttt{PIXEL}), and 1 character-level model (\texttt{CANINE}). First, we propose a scoring Language Quotient (LQ) metric capable of providing a weighted representation of both zero-shot and few-shot evaluation combined. Utilizing this metric, we perform experiments comprising 19 source languages and 133 target languages on three tasks (POS tagging, Dependency parsing, and NER). Our analysis reveals that image-based models excel in cross-lingual transfer when languages are closely related and share visually similar scripts. However, for tasks biased toward word meaning (POS, NER), segmentation-based models prove to be superior. Furthermore, in dependency parsing tasks where word relationships play a crucial role, models with their character-level focus, outperform others. Finally, we propose a recommendation scheme based on our findings to guide model selection according to task and language requirements.
\footnote{The code for reproducing our results is available here \href{https://github.com/mushfiqur11/tokenfreetransfer}{https://github.com/mushfiqur11/tokenfreetransfer}.}

\end{abstract}

\input{Sections/intro.tex}

\input{Sections/method.tex}

\input{Sections/experimentation.tex}

\input{Sections/result.tex}
\input{Sections/recommendation.tex}

\input{Sections/related_work.tex}

\input{Sections/conclusion.tex}

\bibliography{Bib/anthology,Bib/custom}
\bibliographystyle{Bib/acl_natbib}

\appendix

\input{Appendix/Appendix}



\end{document}

%% file: Sections/intro.tex
\section{Introduction}

The performance of multilingual language models varies substantially across languages, with low-resource languages demonstrating particularly sub-optimal results compared to their high-resource counterparts. This disparity poses a global challenge for deploying effective NLP applications, given the diverse linguistic landscape worldwide~\cite{blasi-etal-2022-systematic}.

To address this challenge, cross-lingual transfer has emerged as a promising solution. By leveraging knowledge from high-resource languages, cross-lingual transfer aims to enhance the performance of low-resource ones. However, the effectiveness of cross-lingual knowledge transfer is not uniformly observed across all language pairs. It is influenced by various factors, including language style, structure, origin, dataset quality \cite{yu2022counting, Kreutzer_2022}, and the specific relationship between the source and target languages~\cite{ahmad2019difficulties,he2019cross}. On top of that, the selection of an appropriate language model becomes crucial to achieve successful cross-lingual knowledge transfer. While most state-of-the-art models rely on tokenization~\cite{6289079, 10.5555/177910.177914}, yielding high scores for various linguistic downstream tasks, their performance in terms of cross-lingual transfer has room for further investigation. Considering that word formation can significantly vary across different languages, differences in tokenization techniques can hinder the transfer of linguistic capabilities between languages \cite{ hofmann-etal-2022-embarrassingly}. Hence, the exploration of tokenization-free models is also imperative.

This study thoroughly investigates the role and effectiveness of both tokenization-based~\cite{devlin-etal-2019-bert} and tokenization-free models~\cite{https://doi.org/10.48550/arxiv.2207.06991} in cross-lingual knowledge transfer. Our selection of models encompasses \texttt{BERT} and \texttt{mBERT}~\cite{devlin-etal-2019-bert}, which uses traditional subword-based segmentation. In addition, we delve into tokenization-free models such as \texttt{CANINE}~\cite{Clark_2022} and \texttt{PIXEL}~\cite{https://doi.org/10.48550/arxiv.2207.06991}. \texttt{CANINE} leverages character-level information to accommodate the diverse word formations and structures found in different languages. On the other hand, \texttt{PIXEL} represents texts using visual elements, introducing new possibilities for script-based transfer in visually similar languages.

In this study, we perform standard syntactic task evaluation in both zero-shot and few-shot manner to evaluate the cross-lingual transfer capabilities of these models. While accuracy, F1 score, Labeled Attachment Score (LAS), etc. are all effective evaluation indicators of the goodness of a model, they are not particularly representative of how much a model has learned in a short span of training. We utilize these common metrics over zero-shot and few-shot steps and propose the Learning Quotient (LQ) metric, a novel scoring metric that depends on the relation between the zero-shot and few-shot scores. The metric evaluates the linguistic characteristics of the languages with the model's performance on the tasks. This metric enables a comprehensive evaluation of cross-lingual transfer capabilities, offering valuable insights into the strengths and weaknesses of the models. Our findings suggest contrastive downstream performance that relates to the model architecture. Furthermore, we present a decision tree framework, based on this extensive analysis providing practical guidance for selecting appropriate models based on specific task requirements and language relationships. This framework serves as a tool for researchers and practitioners seeking to harness the potential of NLP applications across diverse languages.

%% file: Sections/method.tex
\section{Methodology}

\paragraph{Problem formulation}

In this work, we use pre-trained language models and fine-tune them on source languages followed by few-shot training on the target languages. Consider the sets of target $T = \{t_1, t_2, \dots, t_m\}$ and source languages $S = \{s_1, s_2, \dots, s_n\}$. We assume source languages $s \in S$ have adequate resources for effective language model training. Conversely, target languages $t \in T$ are low-resource languages with limited data. For any language pair $(s, t)$, we aim to quantify how efficiently a language model can learn the target language $t$ using knowledge transferred from the source language $s$. Given the scarcity of data for $t$, our focus lies on the model's performance in the early stages of fine-tuning it, denoted by the evaluation score $E$.

 Let $(M)_{s}^{\infty}$ represents a language model $M$ fully finetuned on the language $s$ and $(M)_{t}^{c}$ represents the model finetuned up to $c$ steps. We investigate how fast can a model learn the language $t$ in the early steps if it was previously finetuned on $s$. Essentially, we measure the performance of the model $((M)_{s}^{\infty})_{t}^{c}$ where $c$ is a small positive integer. It's important, however, to acknowledge that the efficiency of this method can be influenced by factors such as the similarities between the source and target languages, as well as the quality and quantity of data available for both.

Our methodology can be broadly divided into two steps:


\paragraph{Fine-tuning on Sources} Following the pre-trained model selection, each system is fine-tuned using the selected source languages. This fine-tuning stage allows each system to adjust and optimize its parameters based on specific requirements. Once fine-tuned, the systems are prepared for the evaluation phase in a cross-lingual transfer scenario.

\paragraph{Evaluation and Scoring} The last step involves evaluating each system's performance on target language tasks after undergoing a certain amount of fine-tuning. Two scores are measured at this point: zero-shot and few-shot scores. To measure the final score, we calculate the LQ-score (\S\ref{LQ_metric}). This score allows us to determine the speed and efficiency at which each system learns a new language based on the knowledge transferred from the source language.


\paragraph{Learning Quotient(LQ) metric} \label{LQ_metric}

Let us denote $E_s^{(t_c)}$ as the score achieved by the model $(M){s}^{\infty}$ on the language $t$ after $c$ steps of training on $t$. For different tasks, E can be different. We use accuracy for POS tagging and NER, and Labeled Attachment Score (LAS) for dependency parsing. $E_s^{(t_0)}$ stands for the zero-shot score of the model on $t$. Using the same logic, $\frac{1}{n} \sum_{i=0}^{n} E_{i}^{(t_0)}$ is the average zero-shot score across all source languages, denoted as $Z_A$.

Now, let's introduce our proposed scoring metric, applicable for any pair of languages $t \in T$ and $s \in S$:
\begin{equation}\label{LQ_Score_eqn}
LQ (t,s) = \frac{\left(E_{s}^{(t_c)} - Z_A \right) \left(E_{s}^{(t_c)} + E_{s}^{(t_0)}\right)}{Z_A+ \epsilon}
\end{equation}

$LQ (t,s)$ is comprised of two primary terms, along with a normalization factor. The first term measures the performance of the model after few-shot training on language $t$, relative to the average zero-shot scores for that target language. The second term simply sums the zero-shot and the few-shot scores. To normalize the metric value, we employ the average zero-shot score, $Z_A$. A minute value $\epsilon$ is added to the denominator to avoid division by zero cases. 

The $LQ$ score provides positive reinforcement for both zero-shot and few-shot scores. Any few-shot score that falls below the zero-shot average incurs a substantial penalty. This metric proves effective in quantifying the \textit{pace} at which a model adapts to a new language.\footnote{The proof can be found in Appendix \ref{appendix_lq}}

%% file: Sections/experimentation.tex
\section{Experimentation}

\paragraph{Task Selection} We perform the evaluation on three downstream tasks that heavily depend on fundamental linguistic capabilities and syntactic structure: Dependency Parsing, Part-of-Speech (POS) tagging and Named Entity Recognition (NER). These tasks can work as indicators of a model's understanding of language dynamics and its ability to comprehend and interpret linguistic information \cite{chen2014fast, manning2011part, lample2016neural}

\paragraph{Language and Dataset Selection}

\begin{figure}[t]
    \centering
    \includegraphics[width=\linewidth]{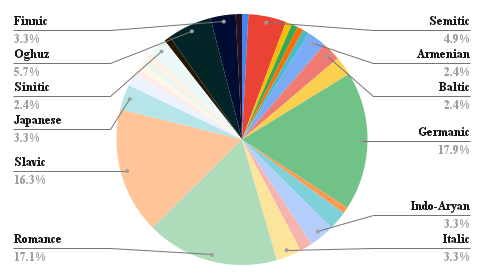}
    \caption{Distribution of the languages according to their sub-families. The majority of these are of Indo-European origin. The languages belong to 28 sub-families spanning 13 different families}
    \label{fig:family_dist}
\end{figure}

\begin{figure}[t]
    \centering
    \includegraphics[width=\linewidth]{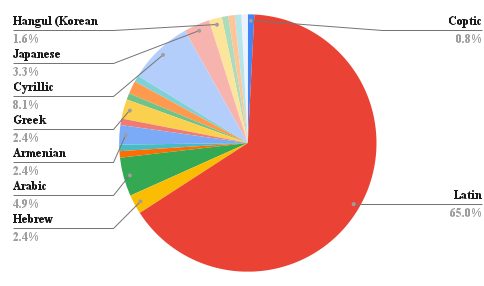}
    \caption{Distribution of the languages according to their scripts. The majority of these use Latin script. The languages use 19 different scripts}
    \label{fig:script_dist}
\end{figure}

\begin{figure*}
\small
    \begin{tabular}{p{.6\textwidth}p{.5\textwidth}}
      \includegraphics[width=\linewidth]{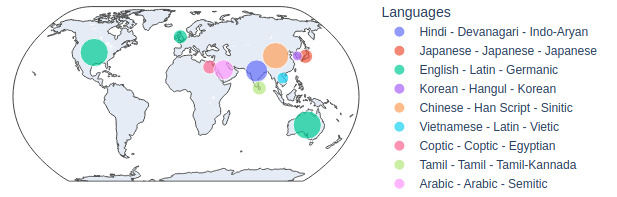}   & 
      \includegraphics[width=.75\linewidth]{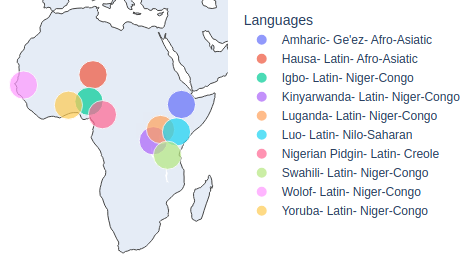} \\
        (a) POS tagging and Dependency Parsing tasks & (b) Named Entity Recognition\\
    \end{tabular}
    \caption{Geographic distribution of source languages (with script and family) used in the analysis across tasks.}
    \label{fig:lang_map}
\end{figure*}

For the execution of POS tagging and Dependency Parsing, we utilized the Universal Dependencies (UD) Dataset ~\cite{nivre-etal-2017-universal,nivre-etal-2020-universal}. To maintain focus and ensure a meaningful study, we selected 9 languages (as listed in Figure \ref{fig:lang_map}(a)) as our source languages and 123 languages as our target languages for the experiments\footnote{A detailed list is provided in appendix \ref{target_lang_list}}. All the models were comprehensively fine-tuned on the selected source languages, thereby establishing a baseline for performance comparison\footnote{All fine-tuned models are available on HuggingFace for further research and investigation}. For NER, we utilized the MashakhaNER dataset ~\cite{adelani-etal-2021-masakhaner} and all its associated languages as sources and targets (as described in Figure \ref{fig:lang_map}(b)). MasakhaNER mainly focuses on a few African languages. These languages are quite low-resource. Hence, these were perfect for this research.

\paragraph{Model Selection} 

To ensure a fair comparison, we use \texttt{BERT},  \texttt{mBERT}, \texttt{CANINE}, and \texttt{PIXEL} as our choice of pre-trained models. \texttt{BERT} and  \texttt{mBERT} use subword segmentation whereas \texttt{CANINE} is a character-based model. Unlike these, \texttt{PIXEL} represents text using visual elements rather than traditional tokens. We selected \texttt{BERT}, as it is the most well-established tokenization-based model that aligns with \texttt{PIXEL}'s pre-training dataset. On the other hand, character-level models provide another perspective for understanding and processing languages, capturing the distinct attributes of word formations. \texttt{CANINE}, with its pre-training on 104 languages, emerged as a strong candidate. As a counterpart, we chose  \texttt{mBERT}, which shares a similar scope of pre-training languages.


\paragraph{Experimental Setup} Our experiments involved two major training phases followed by a result extraction step. In the first training phase, each language model was fully fine-tuned on each of the source languages for each task. 
The experimental setup maintained a high computational standard to ensure robust training and evaluation. All experiments were conducted on 
a remote server equipped with an A100 GPU. The analysis was conducted over 4 (models) x 9 (source languages) x 123 (target languages) data points for Dependency Parsing and POS tagging. For NER, the analysis was conducted over all 4 (models) x 12 (source languages) x 12 (target languages) data points. We used 10 fine-tuning steps (for \S\ref{LQ_Score_eqn}, set $c=10$) for the target languages for all tasks.

For reproducing the results, the language models can be fully fine-tuned on the source languages (our finetuned versions can be used directly from HuggingFace) to get the zero-shot results. These models can then be finetuned on the target languages for 10 steps to get the few-shot score.



%% file: Sections/result.tex
\section{Results and Discussion}

First, we break down the results by several key variables including the visual similarity of languages, their lexical correspondence, and the type of language task. Then, we discuss the performance of these models in light of these variables, revealing patterns regarding model characteristics.

\subsection{Visual similarity is all you need}
\begin{figure*}
    \centering
    \includegraphics[width=.9\textwidth]{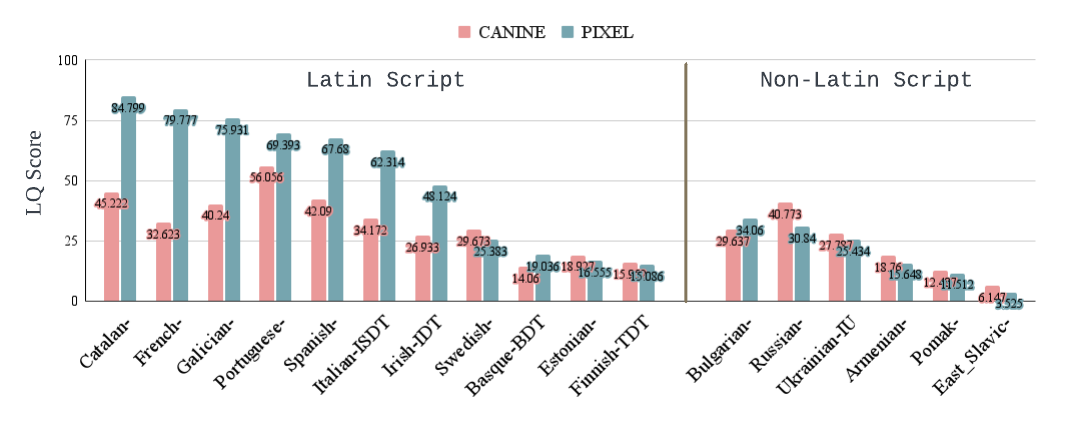}   
    \vspace{-1em}
\caption{LQ score obtained by \texttt{PIXEL} and  \texttt{CANINE} on Latin and non-Latin scripts on POS tagging. \texttt{PIXEL} outperforms  \texttt{CANINE} on the POS tagging task when both source and target use the same script (on the left portion of the graph). Conversely, \texttt{PIXEL} does not outperform  \texttt{CANINE} when the scripts are dissimilar (on the right portion of the graph)}
\label{fig_english_as_source}
\end{figure*}

\paragraph{Case1 (English $\rightarrow$ European)} \label{PIXEL1}


Both of \texttt{PIXEL} and \texttt{BERT} are pre-trained in English. Therefore, for a fair comparison with other models, we perform a comparison where English is the only source language. For evaluation, we consider various European languages, taking into account both lexical similarity and the LQ score on the POS tagging task. Figure \ref{fig_english_as_source} represent the LQ scores of \texttt{PIXEL} and  \texttt{CANINE} when English is used as the source language and various other languages as the targets. Here, in Figure \ref{fig_english_as_source}(a) we observe the proficiency of \texttt{PIXEL} in handling tasks between languages sharing a similar script. For example, English shares similar degrees of lexical similarity with French (0.27) and Russian (0.24) (\S\ref{target_lang_list} and \S\ref{appendix_lex}). However, when considering the LQ scores, French significantly outperforms Russian for \texttt{PIXEL}. Moreover, despite Spanish and Portuguese exhibiting low lexical similarity coefficients with English, they both have achieved high LQ scores. A key factor contributing to these scores is the usage of the Latin script. French, Spanish, and Portuguese, which have all garnered high scores, also use the Latin script. Russian employs a different (Cyrillic) script, which likely explains its relatively lower score. Finnish, despite its use of the Latin script, belongs to a different language family compared to English, which may account for the less impressive performances. Moreover, when the script is non-Latin as presented in Figure~\ref{fig_english_as_source}(b),  \texttt{CANINE} has an edge over \texttt{PIXEL}. 
The lexical similarities between different European languages are outlined in Table \ref{tab_lex_sim} in the appendix.

\paragraph{Case2 (Hindi $\rightarrow$ Urdu | Marathi)}

\begin{table}[ht!]
\centering
\small
\begin{tabular}{l|ll|ll}
\toprule
\multicolumn{5}{c}{POS Tagging}\\\midrule
       & \multicolumn{2}{c|}{Hindi$\rightarrow$Urdu} & \multicolumn{2}{c}{Hindi$\rightarrow$Marathi} \\
   Model    & Score        & Rank       & Score         & Rank        \\ 
       \midrule
PIXEL  & -0.4         & 94         & 17.9          & 5           \\
 \texttt{CANINE} & 96.1         & 3          & 14.6          & 15          \\
 \texttt{mBERT}  & 102.2        & 2          & 7.3           & 112         \\ 
\bottomrule
\end{tabular}
\caption{Comparison between different language models on Hindi as the source and Urdu and Marathi as target shows  \texttt{CANINE} and  \texttt{mBERT} massively favor linguistically similar languages. \texttt{PIXEL} favors visual similarity}
\label{Hindi_as_source}
\end{table}

Despite the high mutual intelligibility and substantial grammatical and linguistic similarities between Hindi and Urdu, as acknowledged in the literature \cite{bhatt2005long}, the LQ score on the POS tagging task attained by \texttt{PIXEL} for this language pairing is not as high as one would anticipate (ranked 94th). The relatively low performance can be attributed to their disparate scripts, underscoring the importance of visual similarity when using image-based language models such as \texttt{PIXEL}. However, for the other three models, with Hindi as the source, Urdu ranked in the top 3 target languages. Table \ref{Hindi_as_source} represents this phenomenon. 

On the flip side, Hindi and Marathi are not mutually intelligible. But both of these languages use the Devanagari script. Sorting the LQ scores for Hindi as the source language, Marathi comes out as one of the top-performing target languages (4th).



\paragraph{Case3 (Arabic $\rightarrow$ X)} 

\begin{table}[t]
    \centering
    \small
    \setlength{\tabcolsep}{3pt}
    \begin{tabular}{l|cccc}
    \toprule
    \multicolumn{5}{c}{Arabic$\rightarrow$X  (POS Tagging)} \\ \midrule
    Lang. (X) & \begin{tabular}[c]{@{}c@{}} \texttt{CANINE}\\LQ Score,\\ (Rank)\end{tabular} & \begin{tabular}[c]{@{}c@{}}\texttt{PIXEL}\\LQ Score,\\ (Rank)\end{tabular} & \begin{tabular}[c]{@{}c@{}}Script\\ Similarity\end{tabular} & \begin{tabular}[c]{@{}c@{}}Linguistics\\ Similarity\end{tabular} \\ \midrule
\multicolumn{1}{l|}{Maltese} & 5.9 (24) & 1.5 (80)   & Dissimilar                                                  & Very Close                                                       \\
\multicolumn{1}{l|}{Persian} & 15.7 (6) & 42.8 (2)    & Same                                                        & Dissimilar                                                       \\
\multicolumn{1}{l|}{Hebrew}  & 43.1 (3) & 36.9 (3)    & Close                                                       & Related                                                   \\
\multicolumn{1}{l|}{Urdu}    & 0.3 (74) & 24.1 (6)    & Same                                                        & Dissimilar \\           
\bottomrule
\end{tabular}    

    \caption{LQ score and rank of \texttt{PIXEL} with Arabic as the source language shows \texttt{PIXEL} receives a high score when scripts are visually similar rather than when languages are only linguistically similar.}
    \label{Arabic_as_source}
\end{table}

In the case of Arabic as the source language, \texttt{PIXEL} received the highest scores for Persian (ranked 2nd) and Urdu (ranked 3rd) as respective source languages. Persian and Urdu are both Indo-European languages and are not at all lexically similar to Arabic. However, these are both written using Arabic script. On the contrary, like Arabic, Maltese is an Afro-Asiatic language with Semitic origin. But \texttt{PIXEL} performed extremely poorly in the case of Maltese (ranked 81st). This, we suspect, is due to the use of Latin script in Maltese, which further emphasizes the effect of visual similarity for \texttt{PIXEL}.

In the case of  \texttt{mBERT} and  \texttt{CANINE}, these patterns of favoring similar-looking scripts were absent. Rather, we saw an average score for the languages irrespective of the script. 
\begin{figure*}
    \centering
    \begin{tabular}{c|c}
     \includegraphics[width=.47\textwidth]{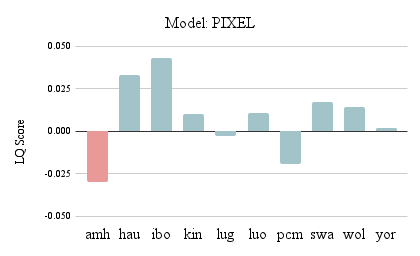}    &  
     \includegraphics[width=.47\textwidth]{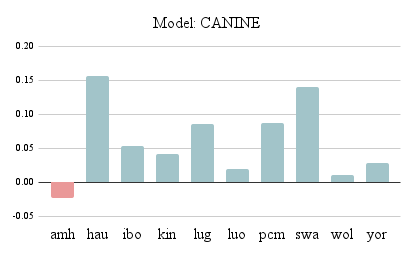} \\
    \end{tabular}
    \caption{Average LQ scores with each language as sources for NER task (for \texttt{PIXEL} and  \texttt{CANINE}) shows Amharic (only non-Latin script) pairs significantly worse with other languages that use Latin script}
    \label{ner_figure}
\end{figure*}
\paragraph{Case4 (African $\rightarrow$ African)}
We've compared all four models using 10 African languages from the MasakhaNER dataset for the Named Entity Recognition (NER) task. Aside from Amharic, which uses the Ge'ez script, all other languages use the Latin script. Figure ~\ref{ner_figure} shows the average LQ score obtained by \texttt{PIXEL} and  \texttt{CANINE} models for each language as sources. The Table shows Amharic as an unfit choice for the source language when the target languages are in Latin script. Comparing \texttt{PIXEL} and  \texttt{CANINE}, we notice  \texttt{CANINE} outperforms \texttt{PIXEL}. Since \texttt{PIXEL} was only pre-trained on English, it is comparatively difficult for \texttt{PIXEL} to perform well on African languages. Conversely,  \texttt{CANINE} was pre-trained on Yoruba (an African language) which has strong linguistic similarities with other African languages. 


\paragraph{Observation} Clearly, the above findings highlight the positive correlation between the performance of \texttt{PIXEL}, an image-based language model, and the visual similarity between languages. It is logical to expect that visually similar language would demonstrate better performance in cross-lingual transfer when utilizing \texttt{PIXEL}. The findings in the  \texttt{CANINE} and  \texttt{mBERT} comparison further reinforce the notion that language models that do not rely on visual representations do not exhibit a strong correlation between their scores and the visual similarity of the source and target languages.


\subsection{Task Specific Performance}
\paragraph{POS tagging} In general, \texttt{mBERT} learns quickly compared to  other models. This can be attributed to several reasons.
First of all, \texttt{mBERT} operates on token-level representations and manifests heavy reliance on word-level semantics. So it is easier to associate the word or subword tokens with their respective POS tags, compared to character-level models like  \texttt{CANINE}.
Moreover, \texttt{mBERT}'s predefined vocabulary, which includes commonly used subwords can potentially expedite the learning process as the model can leverage semantic associations between these known tokens and their POS tags. On the contrary, character-level models have larger input sequence lengths and may require more examples to adequately learn the pattern in data which can lead to slower learning as compared to the tokenization-based models.

In addition, \texttt{mBERT} is trained on multilingual data. So it is more efficient than \texttt{BERT} at transferring knowledge from a high-resource language to a low-resource language, enhancing its few-shot learning capabilities for POS tagging tasks across different languages.

\paragraph{Dependency Parsing} Interestingly, \texttt{CANINE} performs better than  \texttt{mBERT} or \texttt{BERT}.
This may be partly attributed to the nature of the task. Parsing is centered more on understanding the syntactic relationships between words in a sentence rather than on the meanings of individual words. As  \texttt{CANINE} works on character level, it is more equipped to capture finer-grained patterns in these relationships, outperforming  \texttt{mBERT}, exactly because the necessary information is marked with affixal morphemes in many languages. Moreover,  \texttt{CANINE} operates without a predefined vocabulary, and its language independence might be advantageous when parsing sentences in a low-resource language or multilingual context. As a result, it can transfer knowledge across languages more fluidly. On top of that, the occurrence of out-of-vocabulary words or rare words can impact the parsing accuracy. As a character-level model,  \texttt{CANINE} is better equipped in handling out-of-vocabulary words, which might be the reason for its improved performance in parsing in few-shot scenarios. 

\paragraph{Named Entity Recognition} NER, like POS tagging, leans heavily on understanding the meanings of individual words in order to accurately identify and classify named entities. This semantic nature of the task presents an advantage for segmentation-based models such as \texttt{mBERT} over character-level models like \texttt{CANINE}. Despite the multilingual strength of \texttt{CANINE}, its focus on character-level patterns may not sufficiently capture the semantic nuances needed for effective NER. Conversely, \texttt{mBERT}, with its token-based approach, can better handle the word meanings central to NER tasks. Therefore, in our analysis, \texttt{mBERT} demonstrates slightly superior performance in NER compared to \texttt{CANINE}. This suggests that while character-level models may excel in tasks centered on syntactic relationships, segmentation-based models may still hold the edge in tasks with a strong semantic dependency.

\subsection{Unseen Languages}
\begin{table}[t]
\centering
\small
\begin{tabular}{l|c|c|c}
\toprule
\multicolumn{4}{c}{Coptic$\rightarrow$X (POS tagging)}\\\midrule
\textbf{Lang. (X)} & \textbf{ \texttt{mBERT}} & \textbf{ \texttt{CANINE}} & \textbf{BERT} \\\midrule

Telegu & 38.84 & 37.45 & 55.76 \\
French & 20.73 & 26.93 & 50.59 \\
Italian & 22.63 & 26.07 & 47.12 \\
Russian& 33.48 & 27.15 & 43.55 \\
Persian Seraji & 23.21 & 21.26 & 43.53 \\
\bottomrule
\end{tabular}
\caption{Few-shot accuracy for POS tagging task with Coptic as the source language highlighting the performance of \texttt{BERT} (monolingually pre-trained) over  \texttt{mBERT} and  \texttt{CANINE}. Coptic is the only source language (in our analysis) that is not part of the pre-training languages of  \texttt{mBERT} and  \texttt{CANINE} and the only language where \texttt{BERT} significantly outperforms  \texttt{mBERT} and  \texttt{CANINE}}
\vspace{-1.3em}
\label{coptic_as_source}
\end{table}

\texttt{BERT} performs better than  \texttt{mBERT} and  \texttt{CANINE} on some languages that these multilingual models were not pre-trained on. For example, consider the case study of Coptic. In comparison to  \texttt{CANINE} and  \texttt{mBERT}, \texttt{BERT} has better scores for POS tagging when Coptic is used as the source language (Table ~\ref{coptic_as_source}). Multilingual models like  \texttt{CANINE} and  \texttt{mBERT} underperform in this case. Among all the source languages used in our analysis, Coptic is the only source that is not part of the pre-training languages of  \texttt{CANINE} and  \texttt{mBERT}. It is also the only language where \texttt{BERT} has consistently outperformed the multi-lingually pre-trained models.

This inability to effectively adapt to a new unseen language could be attributed to the influence of the scripts of those languages. In these cases, transliterating the target to a high-resource language has been shown to improve performance on downstream tasks~\cite{muller-etal-2021-unseen}. 

%% file: Sections/recommendation.tex
\section{Model Recommendation Tree}

\begin{figure}[t]
    \centering
    \includegraphics[width=\linewidth]{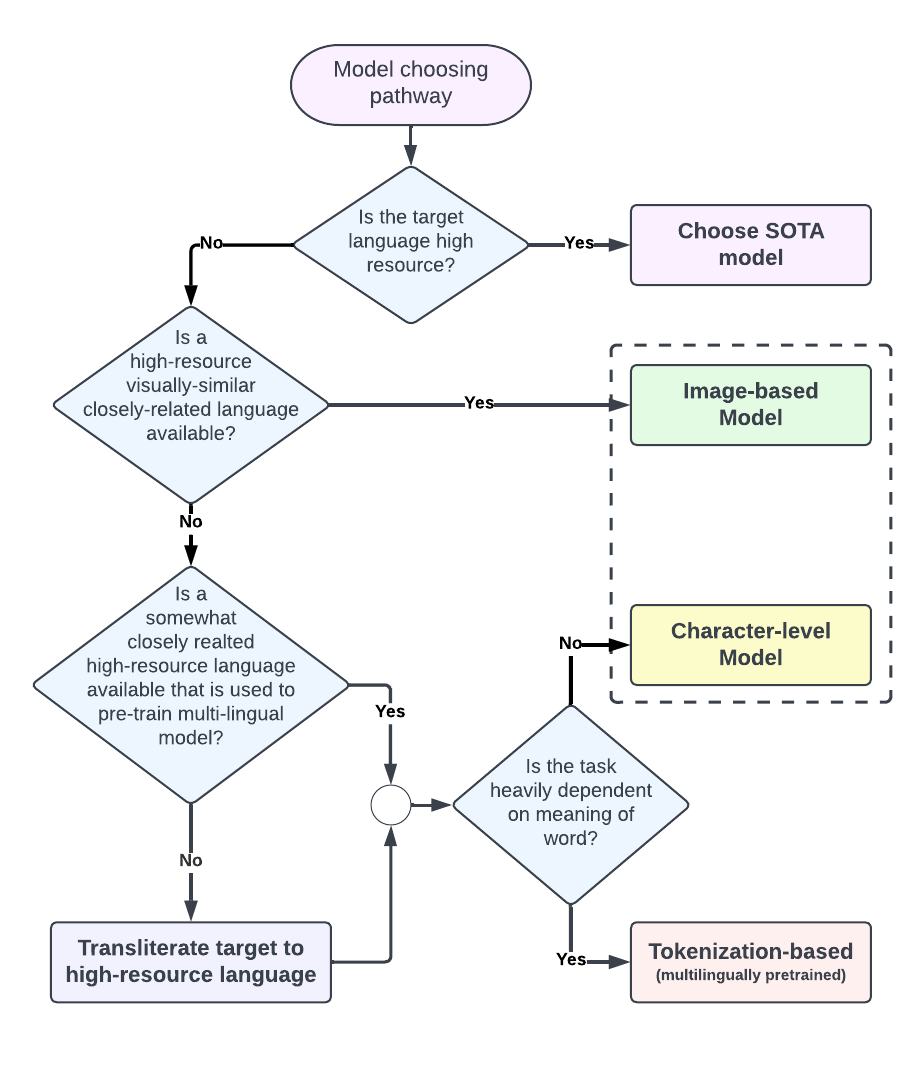}
    \vspace{-1em}
    \caption{Model Recommendation Tree}
    \label{fig:enter-label}
    \vspace{-1em}
\end{figure}







Based on our findings, we propose a model selection pathway predicated on three primary considerations: resource availability for the target language, the presence of a visually similar high-resource language, and the task's semantic dependency.


\paragraph{High Resource Languages}

In the context of high-resource languages, we recommend employing the most advanced models. Our research indicates that both character-based models like \texttt{CANINE} and tokenization-based models like mBERT exhibit superior performances in this setting. Generally, multilingual pre-training grants these models a notable edge over their monolingually trained counterparts, making them well-suited for tasks involving high-resource languages and ensuring efficient performance.

\paragraph{Visual Similarity}

In cases where the target language is resource-poor but visually resembles a high-resource language, our suggestion is to undertake a cross-lingual transfer from the high-resource language using a tokenization-free model like the \texttt{PIXEL}. \texttt{PIXEL} is explicitly designed to discern and capitalize on visual correspondences between languages, which makes it an optimal choice in instances where such resemblances can be exploited.

\paragraph{Semantic Dependency}

If a high-resource language somewhat closely related to the target language has been used in pre-training a multilingual model, the choice between different models should be guided by the task's semantic content requirements. If the task depends heavily on semantic understanding, models like mBERT or similar tokenization-based models are advisable. These models excel in scenarios where deep semantic comprehension is key. Conversely, if the task doesn't require a strong understanding of semantics, character-based models like \texttt{\texttt{CANINE}} may be a more efficient choice. These models typically perform well in scenarios where semantic dependence is lower.

\paragraph{Special Cases}

For scenarios that do not fall within the purview of the above-mentioned conditions, a multitude of factors come into play. For instance, when the source language was not part of the pre-training set for the multilingual model, we suggest transliterating the target language to a high-resource language. Transliterating those languages substantially enhances the performance of these multilingual models on downstream tasks.

%% file: Sections/related_work.tex
\section{Related Work}

\paragraph{Cross-lingual transfer} Cross-lingual transfer has emerged as a valuable approach to enhance model performance in low-resource languages without requiring extensive amounts of target language data \cite{conneau-etal-2020-unsupervised}. XLM-R, proposed by \citeauthor{conneau-etal-2020-unsupervised}, demonstrates the effectiveness of pre-training on a large-scale masked language model trained on 100 languages from CommonCrawl data. It outperforms multilingual BERT (mBERT) on various cross-lingual benchmarks. Similarly, \citeauthor{devlin-etal-2019-bert} and \citeauthor{xue-etal-2021-mt5} propose finetuning approaches for existing pre-trained language models (PLMs). Recently, another approach by \citeauthor{lee-etal-2022-fad} employs adapters for cross-lingual transfer in low-resource languages. Fusing Multiple Adapters for Cross-Lingual Transfer (FAD-X) utilizes language adapters and task adapters to address the imbalance in lower-resource languages. MAD-X \cite{DBLP:journals/corr/abs-2005-00052} is another adapter-based method that employs language, task, and invertible adapters. Moreover, this similar setting coupled with language phylogeny information proved to be useful for low-resource cross-lingual transfer~\cite{faisal-anastasopoulos-2022-phylogeny}.


\paragraph{Tokenization-free models} Tokenization-based models such as \textbf{\texttt{BERT}}~\cite{devlin2019bert}, RoBERTa~\cite{liu2019roberta}, GPT-3~\cite{brown2020language}, ALBERT~\cite{lan2020albert}, T5~\cite{raffel2020exploring} and ELECTRA~\cite{clark2020electra} are leading the field when it comes to performance across a broad range of natural language processing tasks.
However, tokenization-based models like \texttt{BERT} demonstrate poor performance in unexplored domains \cite{https://doi.org/10.48550/arxiv.2010.10392} and lack resilience to noisy data such as typos and missed clicks \cite{https://doi.org/10.48550/arxiv.2003.04985}.

Studies have shown that models using visual text representations are more robust \cite{https://doi.org/10.48550/arxiv.2104.08211}. \texttt{PIXEL} \cite{https://doi.org/10.48550/arxiv.2207.06991} proposes the use of visual embeddings for language modeling, eliminating the need for a fixed vocabulary. Research suggests that models utilizing visual text representations exhibit greater resilience to noisy texts and enable rapid adaptation to new languages while maintaining performance.

\texttt{CANINE} \cite{Clark_2022}, a character-based model, provides an alternative approach that eliminates the reliance on predefined vocabularies. \texttt{CANINE} surpasses vanilla \texttt{BERT} on the TyDiQA benchmark \cite{https://doi.org/10.48550/arxiv.2003.05002} by downsampling input sequences to achieve similar speeds.

ByT5 \cite{https://doi.org/10.48550/arxiv.2105.13626} introduces a modified version of the standard transformer that processes byte sequences, addressing the limitations of a finite vocabulary. Similarly, CHARFORMER \cite{https://doi.org/10.48550/arxiv.2106.12672} proposes a gradient-based sub-word tokenization method that operates directly on a byte level. It performs on par with tokenizer-based approaches and outperforms most byte-level methods.

\paragraph{Language Similarity Metrics}

Several researchers have proposed different methodologies to quantify similarity among languages. For instance, \cite{Petroni_2010} introduced a measure of lexical distance, which quantifies the difference between languages based on their vocabulary. On the other hand, \cite{doi:10.1080/14790710508668395} suggests a metric of linguistic distance that represents how challenging it is for English speakers to learn other languages. However, this method relies on English speakers' learning difficulty, making it language-biased and not generalizable for speakers of other languages.


A different approach is presented by \citeauthor{ciobanu-dinu-2014-automatic}, who propose an automated method for identifying pairs of cognates (words with a common etymology) across languages. But this cognate identification method requires a known list of cognates, limiting its usefulness for less-studied languages, and it may overlook non-lexical aspects of language similarity. 

Another common tool is the Automated Similarity Judgment Program \cite{wiki:wikiasjp} which uses a comprehensive database of vocabulary to analyze linguistic relationships but has been criticized for its simplified standard orthography and its reliance on a limited vocabulary list.

%% file: Sections/conclusion.tex
\section{Conclusion}

This study provides pivotal insights into the practical application of tokenization-based as well as tokenization-free models in cross-lingual transfer tasks, accentuating the importance of context and task-based model selection. However, there's an abundance of uncharted territory awaiting exploration. The gaps in our understanding of tokenization-free models such as \texttt{PIXEL} and \texttt{CANINE} present a significant opportunity for further research. These models, though promising, are still in their early stages of development. This paves the way for studies aiming to enhance their performance, potentially through the integration of advanced learning algorithms or novel feature extraction techniques. 

Additionally, investigating the role of tokenization in handling different language families could provide profound insights. For instance, how do these models perform with agglutinative languages like Turkish or Finnish, or with logographic languages like Chinese? Exploring such linguistic diversity could further clarify the strengths and weaknesses of different model types. An iterative inclusion of extinct or less commonly spoken languages is also essential at this point. 


In summary, this study marks a significant step in understanding the capabilities and limitations of different models in cross-lingual transfer tasks. It opens several doors for future research, promising an exciting trajectory for the evolution of language modeling and translation tasks. The journey ahead, albeit challenging, presents a wealth of opportunities for innovation and discovery.


\section*{Limitations}  This research, while extensive, presents certain limitations. Our study focuses primarily on syntactic tasks, leaving semantic tasks unexplored. While our work delves into the performance of specific models like BERT, mBERT, PIXEL, and CANINE, other models, especially emerging ones like decoder-based language models, remain unexamined in this context. The research also predominantly concerns low-resource languages, potentially limiting the applicability of our findings to high-resource contexts. Moreover, the consideration of different language families, such as agglutinative or logographic languages, is lacking in this analysis. Looking ahead, we plan to address these limitations by incorporating a broader range of language tasks, investigating a wider array of language models, and expanding our research to include high-resource languages and different language families. This will allow us to present a more holistic understanding of cross-lingual transfer in future studies.

\section*{Acknowledgements} We are thankful to the anonymous reviewers for their constructive feedback. Fahim Faisal and Antonios Anastasopoulos are generously supported by the National Science Foundation through grant IIS-2125466.

%% file: Appendix/Appendix.tex
+\section{Appendix}

\subsection{Frequently Asked Questions}
\begin{enumerate}
    \item Q: What did the authors mean by `few-shot' and `zero-shot'?\\
    A: The term `few-shot' is quite loosely used in this paper. Each model is at first fully trained on a source language and then evaluated on some target language. In the evaluation phase, the model is either (i) directly evaluated on the target language (termed as zero-shot), or (ii) fine-tuned for a few steps on the target language (termed as few-shot). 
    \item Q: How can LQ score be negative and what does it imply?\\
    A: The LQ score does not have strong bounds. So it can have negative scores. Since it is a relative metric rather than an absolute one, having a negative score does not create any issue. It implies that the model is performing worse for the source-target pair compared to other sources in the system.
    \item Q: Can LQ metric be used to compare different models?\\
    A: Yes, LQ metric can be used to compare different models if the same pair of source and target languages are considered.
\end{enumerate}

\subsection{LQ Score}
\label{appendix_lq}

\paragraph{Proof of Effectiveness of LQ Score}
Let $E_{s}^{(t_c)} = F$, $E_{s}^{(t_0)} = Z_0$, and $Z_A = \frac{1}{n}\sum_{i=1}^{n}E_i^{(t_0)}$. We can rewrite the LQ score as:
\begin{equation}\label{eqn_1}
LQ(x,k) = \frac{\left(F - Z_A \right) \left(F + Z_0)\right)}{Z_A+ \epsilon}
\end{equation}

We assume that a score would effectively measure the cross-lingual transfer capabilities if it gets positively rewarded for a higher score after a few shots of training in comparison to other language pairs and in comparison to the state before few-shot training. That means the growth of $F$ from $Z_0$ and the difference of $F$ with $Z_A$ should play a high impact on the score.

Simplifying the right-hand-side of Eqn \ref{LQ_Score_eqn}, we get,

\begin{equation}
\frac{F^2 - F  Z_A  + F Z_0 - Z_A  Z_0 }{Z_A+ \epsilon}
\end{equation}

\begin{equation}
= F \frac{F}{Z_A} - F + F \frac{Z_0}{Z_A} - Z_0
\end{equation}

\begin{equation} \label{eqn_simplified}
= F \left( \frac{F+Z_0}{Z_A} \right) - F \left( 1 + \frac{Z_0}{F} \right)
\end{equation}

In equation \ref{eqn_simplified}, the term $\left(F + Z_0\right)/Z_A$ will be greater than 1 when either $F$ is very large or $Z_0$ is significantly larger than $Z_A$. That means a strong positive score can be obtained when the few-shot score is very high or the leap from zero-shot to few-shot is high. The remaining term  $F \left( 1 + \frac{Z_0}{F} \right)$ ensures the stability of the score. So, if a model learns quickly and gains good accuracy/las in the early steps of training, the LQ score will give out a strong score. If a model achieves a good score in zero-shot learning, it also receives a good LQ score.

\paragraph{Limitations of LQ Score}
The score utilizes a normalizing term that averages the zero-shot scores across all source languages. So, for any pair of languages, $x$ and $k$, the LQ score will not always be the same. It will vastly depend on the list of source languages used in the experimentation. So, the numeric value of the LQ score does not have a direct meaning. However, for a given source, the relation between the target languages is indicative of how compatible the source and target are. On the flip side, for a target language, the relation between the source languages is also meaningful.

\subsection{Hyper-parameters}
\label{sec:parameters}

\subsubsection{Dependency Parsing}
\paragraph{Full Fine-tuning (on source)}
\begin{itemize}[noitemsep,nolistsep,leftmargin=*]
    \item Train batch size: 32
    \item Max Training Steps: 15000
    \item Early Stopping: Yes
    \item Learning Rate: 5e-5
    \item Maximum Sequence Length: 256
    \item Eval metric: LAS
\end{itemize}

\paragraph{Few-shot Fine-tuning (on targets)}
\begin{itemize}[noitemsep,nolistsep,leftmargin=*]
    \item Train batch size: 32
    \item Max Training Steps: 10
    \item Learning Rate: 5e-5
    \item Maximum Sequence Length: 256
    \item Eval metric: LAS
\end{itemize}

\subsubsection{POS Tagging}
\paragraph{Full Fine-tuning (on source)}
\begin{itemize}[noitemsep,nolistsep,leftmargin=*]
    \item Train batch size: 32
    \item Max Training Steps: 15000
    \item Early Stopping: Yes
    \item Learning Rate: 5e-5
    \item Maximum Sequence Length: 256
    \item Eval metric: Accuracy
\end{itemize}

\paragraph{Few-shot Fine-tuning (on targets)}
\begin{itemize}[noitemsep,nolistsep,leftmargin=*]
    \item Train batch size: 32
    \item Max Training Steps: 10
    \item Learning Rate: 5e-5
    \item Maximum Sequence Length: 256
    \item Eval metric: Accuracy
\end{itemize}

\subsubsection{Named Entity Recognition}
\paragraph{Full Fine-tuning (on source)}
\begin{itemize}[noitemsep,nolistsep,leftmargin=*]
    \item Train batch size: 32
    \item Max Training Steps: 15000
    \item Early Stopping: Yes
    \item Learning Rate: 5e-5
    \item Maximum Sequence Length: 256
    \item Eval metric: Accuracy
\end{itemize}
\paragraph{Few-shot Fine-tuning (on targets)}
\begin{itemize}[noitemsep,nolistsep,leftmargin=*]
    \item Train batch size: 32
    \item Max Training Steps: 10
    \item Learning Rate: 5e-5
    \item Maximum Sequence Length: 256
    \item Eval metric: Accuracy
\end{itemize}

\subsection{Source languages as target languages}
Table 4 provides a comprehensive analysis of the PIXEL model's performance in terms of accuracy in the POS-tagging task, evaluated in both zero-shot and few-shot scenarios. Here, the set of source languages also serves as the target languages, creating a self-referential evaluation method. This unique approach further allows for a deeper understanding of the model's strengths and weaknesses when dealing with identical sources and target languages.

\input{Appendix/pixel_pos}

\subsection{List of target languages} \label{target_lang_list}
\input{Appendix/target_language_list}
Tables \ref{target_list_1}, \ref{target_list_2}, and \ref{target_list_3} give an elaborate list of languages and their scripts along with their respective families. The languages are spread across multiple scripts and multiple families.

\subsection{Lexical Similarity}\label{appendix_lex}

Lexical similarity is the percentage obtained by comparing standardized wordlists from two linguistic varieties and tallying words similar in form and meaning~\cite{wiki:ethnologue}. It ranges from 0 to 100, representing the vocabulary overlap between two languages. Values over 85\% often suggest the speech variant may be a dialect of the compared language. The proportion of lexical similarity between two kinds of language is calculated by comparing standardized lists of words and tallying the forms that demonstrate similarity in both structure and meaning.

\input{Appendix/lexical_similarity}

Table \ref{tab_lex_sim} gives the similarity scores between different European Language pairs \cite{wiki:ethnologue, fan2021discovering}.

\section{Additional Materials}

\begin{table*}[ht!]
\small
\centering
\begin{tabular}{l|l|l|l}
\toprule
                                 & mBERT       & CANINE      & BERT        \\ 
                                 \midrule
UD\_Telugu-MTG                   & 38.83      & 37.45      & 55.76      \\ 
UD\_French-ParTUT                & 20.37      & 26.93       & 50.52      \\ 
UD\_Italian-ParTUT               & 22.63      & 26.07      & 47.12      \\ 
UD\_French-Sequoia               & 22.57      & 27.72      & 46.64      \\ 
UD\_Spanish-AnCora               & 24.10      & 24.17      & 46.09       \\ 
UD\_French-GSD                   & 22.94      & 28.09      & 46.03       \\ 
UD\_Galician-CTG                 & 27.80      & 22.67      & 45.95      \\ 
UD\_Italian-ISDT                 & 23.07      & 26.80      & 45.62      \\ 
UD\_Italian-VIT                  & 24.43      & 27.54      & 44.61       \\ 
UD\_Spanish-GSD                  & 22.55      & 23.2        & 43.80      \\ 
UD\_Russian-GSD                  & 33.48      & 27.15      & 43.54      \\ 
UD\_Persian-Seraji               & 23.21      & 21.26      & 43.54      \\ 
UD\_Catalan-AnCora               & 22.42      & 23.93      & 43.41      \\ 
UD\_Turkish-Kenet                & 32.31      & 32.29      & 43.21      \\ 
UD\_Portuguese-Bosque            & 26.99      & 22.92      & 42.51      \\ 
UD\_Portuguese-GSD               & 26.36      & 22.36      & 41.95      \\ 
UD\_Italian-MarkIT               & 21.57      & 26.19      & 41.78      \\ 
UD\_Turkish-FrameNet             & 33.33      & 32.45      & 41.38      \\ 
UD\_Turkish-Penn                 & 29.87       & 30.68      & 41.25      \\ 
UD\_French-Rhapsodie             & 27.63      & 32.16      & 40.88      \\ 
UD\_Hebrew-IAHLTwiki             & 26.53      & 19.43       & 40.13      \\ 
UD\_Russian-SynTagRus            & 33.16      & 27.29      & 40.09      \\ 
UD\_Polish-PDB                   & 30.01      & 25.15      & 39.90      \\ 
UD\_Lithuanian-ALKSNIS           & 34.08      & 25.40      & 39.78       \\ 
UD\_Arabic-PADT                  & 30.52       & 19.67       & 39.62      \\ 
UD\_Belarusian-HSE               & 30.87      & 23.30      & 38.41      \\ 
UD\_Polish-LFG                   & 30.18      & 29.38      & 38.24       \\ 
UD\_Ukrainian-IU                 & 30.56      &             & 37.60      \\ 
UD\_Hebrew-HTB                   & 23.88      & 17.32      & 37.58      \\ 
UD\_Vietnamese-VTB               & 21.60      & 25.97      & 37.52      \\ 
UD\_Turkish-BOUN                 & 30.42      & 25.66      & 37.35      \\ 
UD\_Greek-GDT                    & 25.18      & 15.39      & 37.26       \\ 
UD\_Latvian-LVTB                 & 32.35      & 24.42      & 37.24      \\ 
UD\_Romanian-SiMoNERo            & 34.12      & 21.87      & 37.23      \\ 
\bottomrule
\end{tabular}

\caption{LQ scores of different models (using Coptic as source language)}
\end{table*}

%% file: Appendix/pixel_pos.tex
\begin{table*}[ht!]

\centering
\small

\begin{tabular}{lccccccccc|c}
\toprule
\multicolumn{1}{c}{\begin{tabular}[c]{@{}c@{}}Target \\ Language\end{tabular}} & \textbf{English} & \textbf{Arabic} & \textbf{Korean} & \textbf{Vietnamese} & \textbf{Tamil} & \textbf{Chinese} & \textbf{Japanese} & \textbf{Coptic} & \textbf{Hindi} & \textbf{Average ($Z_A$}) \\ \midrule
\multicolumn{1}{l|}{English}                                                   & \textbf{0.967}   & 0.238           & 0.297           & 0.284               & 0.255          & 0.149            & 0.297             & 0.289           & 0.219      & 0.33    \\
\multicolumn{1}{l|}{Arabic}                                                    & 0.238            & \textbf{0.958}  & 0.412           & 0.379               & 0.289          & 0.152            & 0.403             & 0.177           & 0.07       & 0.34    \\
\multicolumn{1}{l|}{Korean}                                                    & 0.28             & 0.382           & \textbf{0.944}  & 0.476               & 0.284          & 0.23             & 0.413             & 0.329           & 0.172      & 0.39    \\
\multicolumn{1}{l|}{Vietnamese}                                                & 0.286            & 0.341           & 0.47            & \textbf{0.86}       & 0.3            & 0.234            & 0.458             & 0.321           & 0.233      & 0.39    \\
\multicolumn{1}{l|}{Tamil}                                                     & 0.135            & 0.3             & 0.388           & 0.331               & \textbf{0.817} & 0.224            & 0.37              & 0.25            & 0.223      & 0.34    \\
\multicolumn{1}{l|}{Chinese}                                                   & 0.336            & 0.32            & 0.428           & 0.412               & 0.3            & \textbf{0.93}    & 0.525             & 0.3             & 0.274       & 0.43   \\
\multicolumn{1}{l|}{Japanese}                                                  & 0.276            & 0.294           & 0.376           & 0.349               & 0.229          & 0.303            & \textbf{0.973}    & 0.226           & 0.179       & 0.36   \\
\multicolumn{1}{l|}{Coptic}                                                    & 0.103            & 0.144           & 0.189           & 0.188               & 0.154          & 0.056            & 0.162             & \textbf{0.962}  & 0.093       & 0.23   \\
\multicolumn{1}{l|}{Hindi}                                                     & 0.229            & 0.215           & 0.292           & 0.302               & 0.24           & 0.202            & 0.274             & 0.209           & \textbf{0.964} & 0.33\\
\bottomrule
\end{tabular}

\vspace{0.5cm} 
(a) Accuracy for POS task at zero-shot 
\vspace{0.5cm} \\
\begin{tabular}{lccccccccc}
\toprule
                                & \textbf{Arabic} & \textbf{Chinese} & \textbf{Coptic} & \textbf{English} & \textbf{Hindi} & \textbf{Japanese} & \textbf{Korean} & \textbf{Tamil} & \textbf{Vietnamese} \\ \midrule
\multicolumn{1}{l|}{Arabic}     & \textbf{0.958}  & 0.328            & 0.337           & 0.396            & 0.277          & 0.34              & 0.388           & 0.337          & 0.355               \\
\multicolumn{1}{l|}{Chinese}    & 0.371           & \textbf{0.93}    & 0.339           & 0.366            & 0.395          & 0.531             & 0.414           & 0.328          & 0.391               \\
\multicolumn{1}{l|}{Coptic}     & 0.191           & 0.11             & \textbf{0.962}  & 0.183            & 0.163          & 0.188             & 0.193           & 0.166          & 0.229               \\
\multicolumn{1}{l|}{English}    & 0.25            & 0.219            & 0.324           & \textbf{0.968}   & 0.283          & 0.304             & 0.292           & 0.265          & 0.29                \\
\multicolumn{1}{l|}{Hindi}      & 0.311           & 0.288            & 0.331           & 0.319            & \textbf{0.964} & 0.264             & 0.261           & 0.257          & 0.349               \\
\multicolumn{1}{l|}{Japanese}   & 0.417           & 0.403            & 0.295           & 0.374            & 0.334          & \textbf{0.973}    & 0.385           & 0.295          & 0.364               \\
\multicolumn{1}{l|}{Korean}     & 0.42            & 0.373            & 0.416           & 0.404            & 0.403          & 0.409             & \textbf{0.943}  & 0.384          & 0.47                \\
\multicolumn{1}{l|}{Tamil}      & 0.328           & 0.303            & 0.298           & 0.33             & 0.298          & 0.302             & 0.39            & \textbf{0.817} & 0.337               \\
\multicolumn{1}{l|}{Vietnamese} & 0.385           & 0.312            & 0.328           & 0.379            & 0.395          & 0.439             & 0.454           & 0.336          & \textbf{0.859} \\ 
\bottomrule
\end{tabular}

\vspace{0.5cm} 
(b) Accuracy for POS task at few-shot 
\vspace{0.5cm} \\

\caption{Accuracy of PIXEL model (on POS-tagging task) of zero-shot evaluation and few-shot evaluation of 9 source languages on the same languages as targets}

\label{pixel-pos}
\end{table*}

%% file: Appendix/target_language_list.tex
\begin{table*}[ht!]
\small
\centering
\begin{tabular}{llll}
\toprule
\textbf{Language Name} & \textbf{Script} & \textbf{Language Family} & \textbf{Sub-family} \\ \midrule
Armenian-ArmTDP & Armenian & Indo-European & Armenian \\
Armenian-BSUT & Armenian & Indo-European & Armenian \\
Western\_Armenian-ArmTDP & Armenian & Indo-European & Armenian \\
Latvian-LVTB & Latin & Indo-European & Baltic \\
Lithuanian-ALKSNIS & Latin & Indo-European & Baltic \\
Lithuanian-HSE & Latin & Indo-European & Baltic \\
Irish-IDT & Latin & Indo-European & Celtic \\
Scottish\_Gaelic-ARCOSG & Latin & Indo-European & Celtic \\
Welsh-CCG & Latin & Indo-European & Celtic \\
Afrikaans-AfriBooms & Latin & Indo-European & Germanic \\
Danish-DDT & Latin & Indo-European & Germanic \\
Dutch-Alpino & Latin & Indo-European & Germanic \\
Dutch-LassySmall & Latin & Indo-European & Germanic \\
English-Atis & Latin & Indo-European & Germanic \\
English-ESL & Latin & Indo-European & Germanic \\
English-EWT & Latin & Indo-European & Germanic \\
English-GUM & Latin & Indo-European & Germanic \\
English-GUMReddit & Latin & Indo-European & Germanic \\
English-LinES & Latin & Indo-European & Germanic \\
English-ParTUT & Latin & Indo-European & Germanic \\
Faroese-FarPaHC & Latin & Indo-European & Germanic \\
German-GSD & Latin & Indo-European & Germanic \\
German-HDT & Latin & Indo-European & Germanic \\
Icelandic-IcePaHC & Latin & Indo-European & Germanic \\
Icelandic-Modern & Latin & Indo-European & Germanic \\
Norwegian-Bokmaal & Latin & Indo-European & Germanic \\
Norwegian-Nynorsk & Latin & Indo-European & Germanic \\
Norwegian-NynorskLIA & Latin & Indo-European & Germanic \\
Swedish-LinES & Latin & Indo-European & Germanic \\
Swedish-Talbanken & Latin & Indo-European & Germanic \\
Gothic-PROIEL & Gothic & Indo-European & Germanic \\
Turkish\_German-SAGT & Latin & Indo-European & Germanic (German) \\
Ancient\_Greek-Perseus & Greek & Indo-European & Hellenic \\
Ancient\_Greek-PROIEL & Greek & Indo-European & Hellenic \\
Greek-GDT & Greek & Indo-European & Hellenic \\
Hindi\_English-HIENCS & Devanagari and Latin & Indo-European & Indo-Aryan \\
Hindi-HDTB & Devanagari & Indo-European & Indo-Aryan \\
Marathi-UFAL & Devanagari & Indo-European & Indo-Aryan \\
Urdu-UDTB & Arabic & Indo-European & Indo-Aryan \\
Persian-PerDT & Arabic & Indo-European & Iranian \\
Persian-Seraji & Arabic & Indo-European & Iranian \\
Latin-ITTB & Latin & Indo-European & Italic \\
Latin-LLCT & Latin & Indo-European & Italic\\
\bottomrule
\end{tabular}
\caption{List of Target Languages (Part 1)}
\label{target_list_1}
\end{table*}

\begin{table*}[ht!]
\small
\centering
\begin{tabular}{llll}
\toprule
\textbf{Language Name} & \textbf{Script} & \textbf{Language Family} & \textbf{Sub-family} \\ \midrule
Latin-PROIEL & Latin & Indo-European & Italic \\
Latin-UDante & Latin & Indo-European & Italic \\
Catalan-AnCora & Latin & Indo-European & Romance \\
French-FTB & Latin & Indo-European & Romance \\
French-GSD & Latin & Indo-European & Romance \\
French-ParTUT & Latin & Indo-European & Romance \\
French-Rhapsodie & Latin & Indo-European & Romance \\
French-Sequoia & Latin & Indo-European & Romance \\
Galician-CTG & Latin & Indo-European & Romance \\
Italian-ISDT & Latin & Indo-European & Romance \\
Italian-MarkIT & Latin & Indo-European & Romance \\
Italian-ParTUT & Latin & Indo-European & Romance \\
Italian-PoSTWITA & Latin & Indo-European & Romance \\
Italian-TWITTIRO & Latin & Indo-European & Romance \\
Italian-VIT & Latin & Indo-European & Romance \\
Old\_French-SRCMF & Latin & Indo-European & Romance \\
Portuguese-Bosque & Latin & Indo-European & Romance \\
Portuguese-GSD & Latin & Indo-European & Romance \\
Romanian-Nonstandard & Latin & Indo-European & Romance \\
Romanian-RRT & Latin & Indo-European & Romance \\
Romanian-SiMoNERo & Latin & Indo-European & Romance \\
Spanish-AnCora & Latin & Indo-European & Romance \\
Spanish-GSD & Latin & Indo-European & Romance \\
Croatian-SET & Latin & Indo-European & Slavic \\
Czech-CAC & Latin & Indo-European & Slavic \\
Czech-CLTT & Latin & Indo-European & Slavic \\
Czech-FicTree & Latin & Indo-European & Slavic \\
Czech-PDT & Latin & Indo-European & Slavic \\
Polish-LFG & Latin & Indo-European & Slavic \\
Polish-PDB & Latin & Indo-European & Slavic \\
Slovak-SNK & Latin & Indo-European & Slavic \\
Slovenian-SSJ & Latin & Indo-European & Slavic \\
Old\_Church\_Slavonic-PROIEL & Glagolitic and Cyrillic & Indo-European & Slavic \\
Belarusian-HSE & Cyrillic & Indo-European & Slavic \\
Bulgarian-BTB & Cyrillic & Indo-European & Slavic \\
Old\_East\_Slavic-Birchbark & Cyrillic & Indo-European & Slavic \\
Old\_East\_Slavic-TOROT & Cyrillic & Indo-European & Slavic \\
Pomak-Philotis & Cyrillic & Indo-European & Slavic \\
Russian-GSD & Cyrillic & Indo-European & Slavic \\
Russian-SynTagRus & Cyrillic & Indo-European & Slavic \\
Russian-Taiga & Cyrillic & Indo-European & Slavic \\
Serbian-SET & Cyrillic & Indo-European & Slavic \\
Ukrainian-IU & Cyrillic & Indo-European & Slavic\\
\bottomrule
\end{tabular}

\caption{List of Target Languages (Part 2)}
\label{target_list_2}
\end{table*}

\begin{table*}[ht!]
\small
\centering
\begin{tabular}{llll}\toprule
\textbf{Language Name} & \textbf{Script} & \textbf{Language Family} & \textbf{Sub-family} \\ \midrule
Coptic-Scriptorium & Coptic & Afro-Asiatic & Egyptian \\
Maltese-MUDT & Latin & Afro-Asiatic & Semitic \\
Ancient\_Hebrew-PTNK & Hebrew & Afro-Asiatic & Semitic \\
Hebrew-HTB & Hebrew & Afro-Asiatic & Semitic \\
Hebrew-IAHLTwiki & Hebrew & Afro-Asiatic & Semitic \\
Arabic-NYUAD & Arabic & Afro-Asiatic & Semitic \\
Arabic-PADT & Arabic & Afro-Asiatic & Semitic \\
Vietnamese-VTB & Latin & Austroasiatic & Vietic \\
Indonesian-GSD & Latin & Austronesian & Malayo-Polynesian \\
Tamil-TTB & Tamil & Dravidian & Tamil-Kannada \\
Telugu-MTG & Telugu & Dravidian & Telugu-Kui \\
Japanese-BCCWJ & Japanese (Kanji, Hiragana, Katakana) & Japonic & Japanese \\
Japanese-BCCWJLUW & Japanese (Kanji, Hiragana, Katakana) & Japonic & Japanese \\
Japanese-GSD & Japanese (Kanji, Hiragana, Katakana) & Japonic & Japanese \\
Japanese-GSDLUW & Japanese (Kanji, Hiragana, Katakana) & Japonic & Japanese \\
Korean-GSD & Hangul and Hanja & Koreanic & Korean \\
Korean-Kaist & Hangul and Hanja & Koreanic & Korean \\
Basque-BDT & Latin & Language Isolate & Language Isolate \\
Naija-NSC & Latin & Niger-Congo & Benue-Congo \\
Wolof-WTB & Latin & Niger-Congo & Senegambian \\
Swedish\_Sign\_Language & Swedish Sign Language (SignWriting) & Sign Language & Sign Language \\
Chinese-GSDSimp & Simplified Chinese (Han script) & Sino-Tibetan & Sinitic \\
Classical\_Chinese-Kyoto & Classical Chinese (Han script) & Sino-Tibetan & Sinitic \\
Chinese-GSD & Chinese (Han script) & Sino-Tibetan & Sinitic \\
Uyghur-UDT & Arabic & Turkic & Karluk \\
Turkish-Atis & Latin & Turkic & Oghuz \\
Turkish-BOUN & Latin & Turkic & Oghuz \\
Turkish-FrameNet & Latin & Turkic & Oghuz \\
Turkish-IMST & Latin & Turkic & Oghuz \\
Turkish-Kenet & Latin & Turkic & Oghuz \\
Turkish-Penn & Latin & Turkic & Oghuz \\
Turkish-Tourism & Latin & Turkic & Oghuz \\
Estonian-EDT & Latin & Uralic & Finnic \\
Estonian-EWT & Latin & Uralic & Finnic \\
Finnish-FTB & Latin & Uralic & Finnic \\
Finnish-TDT & Latin & Uralic & Finnic \\
Hungarian-Szeged & Latin & Uralic & Ugric\\
\bottomrule
\end{tabular}

\caption{List of Target Languages (Part 3)}
\label{target_list_3}
\end{table*}

%% file: Appendix/lexical_similarity.tex
\begin{table*}[ht!]

\small

\begin{tabular}{lccccccccc}
\toprule
\multicolumn{1}{c}{} & Catalan & English & French & German & Italian & Portuguese & Romanian & Russian & Spanish \\ \midrule
\multicolumn{1}{l|}{Catalan}            & 1       & -       & 0.85   & -      & 0.87    & 0.85       & 0.73     & -       & 0.85    \\
\multicolumn{1}{l|}{English}            & -       & 1       & 0.27   & 0.6    & -       & -          & -        & 0.24    & -       \\
\multicolumn{1}{l|}{French}             & 0.85    & 0.27    & 1      & 0.29   & 0.89    & 0.75       & 0.75     & -       & 0.75    \\
\multicolumn{1}{l|}{German}             & -       & 0.6     & 0.29   & 1      & -       & -          & -        & -       & -       \\
\multicolumn{1}{l|}{Italian}            & 0.87    & -       & 0.89   & -      & 1       & 0.8        & 0.77     & -       & 0.82    \\
\multicolumn{1}{l|}{Portuguese}         & 0.85    & -       & 0.75   & -      & 0.8     & 1          & 0.72     & -       & 0.89    \\
\multicolumn{1}{l|}{Romanian}           & 0.73    & -       & 0.75   & -      & 0.77    & 0.72       & 1        & -       & 0.71    \\
\multicolumn{1}{l|}{Russian}            & -       & 0.24    & -      & -      & -       & -          & -        & 1       & -       \\
\multicolumn{1}{l|}{Spanish}            & 0.85    & -       & 0.75   & -      & 0.82    & 0.89       & 0.71     & -       & 1\\    
\bottomrule
\end{tabular}

\caption{Lexical similarity among European languages~\cite{wiki:ethnologue, fan2021discovering}}
\label{tab_lex_sim}
\end{table*}